# Yet Unnoticed in LSTM: Binary Tree Based Input Reordering, Weight Regularization, and Gate Nonlinearization

Mojtaba Moattari
Independent Researcher
Moatary.m@gmail.com
Aug 26, 2025

*Abstract*— LSTM models used in current Machine Learning literature and applications, has a promising solution for permitting long term information using gating mechanisms that forget and reduce effect of current input information. However, even with this pipeline, they do not optimally focus on specific old index or long-term information. This paper elaborates upon input reordering approaches to prioritize certain input indices.

Moreover, no LSTM based approach is found in the literature that examines weight normalization while choosing the right weight and exponent of Lp norms through main supervised loss function. In this paper, we find out which norm best finds relationship between weights to either smooth or sparsify them.

Lastly, gates, as weighted representations of inputs and states, which control reduction-extent of current input versus previous inputs (~ state), are not nonlinearized enough (through a small FFNN). As analogous to attention mechanisms, gates easily filter current information to bold (emphasize on) past inputs. Nonlinearized gates can more easily tune up to peculiar nonlinearities of specific input in the past. This type of nonlinearization is not proposed in the literature, to the best of author's knowledge.

The proposed approaches are implemented and compared with a simple LSTM to understand their performance in text classification tasks. The results show they improve accuracy of LSTM.

## I. Introduction

Long Short Term Memory (LSTM) networks, are Seq2Seq models known for ability to preserve, capture and transform long term textual or visuo-linguistic information. LSTM [3, 4], thanks to its input, output, and forget gates, addresses Recurrent Neural Network's gradient vanishing and exploding, and tackles insufficient long term memory by learning the gating parameters to reduce or increase gating that directly affect intensity of hidden states. The effectiveness of LSTMs has been demonstrated at numerous sequence-related tasks such as Text Generation [5]; [6], sequence-to-sequence translation [13], image captions generation [14, 15], and generating source code [16]. LSTMs are still used in highly relevant applications [17, 18] and have stood the test of time.

Table 1. Advantages and disadvantages of hybridizing LSTM with different Deep Learning models

| Name | Inductive Bias | Advantages | Disadvantages |
|---|---|---|---|
| Bidirectional LSTM [10] | Backpropagation through time | Relation to both previous and upcoming words | Not available for image-text data classification |
| CNN_LSTM | Backpropagation through time | CNN selects the important features for LSTM | Not available for only-text data classification |
| Bi-LSTM-RBM [2] | Backpropagation th1rough time and space | Extraction of hierarchical, abstract, and long-term information<br><br>combines deep Bi-LSTM with a Restricted Boltzmann Machine (RBM). This hybrid approach aims to improve the accuracy of object detection | Not available for only-text data classification |
| LSTM-Attention-LSTM [1] | Per each state | This encoder-attention-decoder model overcome disadvantage of coder-decoder model by ability to obtain sufficiently long input sequence<br><br>fixed-length vector cannot fully represent the whole sequence. | Higher computational burden |
| xLSTM [7] | Per each state | xLSTM introduces exponential gating and memory mixing that are beneficial for long term sequen- tial learning. | Higher computational burden |

Yet, there is no research in the literature that does comparative analysis of weight regularization to control the amount of weight-closeness.

Also, information-theoretical inspiration of human brain is not yet entered LSTM designs, making LSTM similar to recursive LSTM (designed by Socher et al.).

Moreover, due to sharedness of parameters through time, only 4 weight matrices has to be optimized, and they have to be sensitive to all time locations during learning process. This makes having more weights necessary due to the following reasons:

- For nonlinear relationships between current unit and next unit, more number of activations and weights matrices are necessary.
- Adding more parameters matrices, provide more plasticity and learnability for the LSTM model.

List of challenges addressed by LSTM:

- ability to handle the exploding / vanishing gradient problem by the art of selecting a fit activation function.
- learning long-term dependencies by the art of using forget gates

Shi et al. proposed Exponential Smoothing for reducing attention parameter weights [8]. Although this regularization can make the model achieve higher accuracy with fewer parameters, a high chance of overfit can threaten the model using this approach.

Contribution of this paper is as follows:

- This paper uses input index Huffman coding (hierarchical binary selection of input index) to elaborate upon input reordering approaches to prioritize more spread input indices later. This makes the later selected inputs have shallower backpropagation to be safe from gradient vanishing and exploding. To the best of author's knowledge, this is the first time input reordering is using ideas of Information Theory.
- We also propose a new weight normalization approach to find out whether spiky parameters or smooth parameters improve accuracy.
- Moreover, we propose a weight nonlinearization approach that introduce new parameter matrices. This increases memory power of whole LSTM. Nonlinearized gates can more easily tune up to peculiar nonlinearities of specific input in the past.

The proposed approaches are implemented and compared with a simple LSTM to understand their performance in text classification tasks. The results show they improve accuracy of LSTM.

The remainder of this paper is organized as follows.

Chapter 2 explains the related works that are important in analyzing and realizing both the limitation of current LSTM, and the crucialness of proposed methods. Chapter 3, the Proposed Methods, explains each of the contributions one by one, with possible outcomes that might happen due to errors in their structure. Chapter 4 provides Hyperparameter Tuning, Baseline model, and the datasets used. Afterwards, the performance comparisons of each proposed models are described and analyzed. Finally, the paper is closed with conclusion and future works.

## II. RELATED WORKS

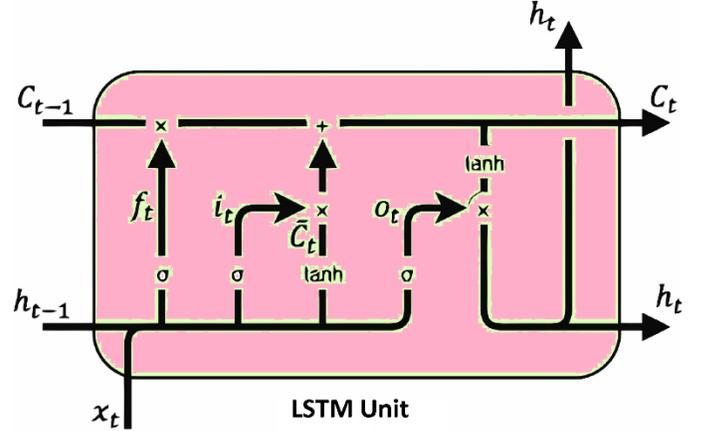

Figure 1. The architecture of a LSTM unit

The architecture of a LSTM unit is shown in Figure 1.

$$i_t = \sigma(x_t U^i + h_{t-1} W^i) \quad (1)$$
$$f_t = \sigma(x_t U^f + h_{t-1} W^f) \quad (2)$$
$$o_t = \sigma(x_t U^o + h_{t-1} W^o) \quad (3)$$
$$\widetilde{C}_t = \tanh(x_t U^g + h_{t-1} W^g) \quad (4)$$
$$C_t = \sigma(f_t * C_{t-1} + i_t * \widetilde{C}_t) \quad (5)$$
$$h_t = \tanh(C_t) * o_t \quad (6)$$

Where $\sigma$ is the Sigmoid activation function; $\odot$ is the Hadamard product, and $x_t$ is the current input vector; $U^f, U^i, U^c, U^o$ are related to input of the corresponding weight vector; $h_{t-1}$ is the state vector of the previous cell unit's hidden layer; $W^f, W^i, W^c, W^o$, respectively, are about the hidden layer weights of state vector.

**Remembering past by forgetting the present.** The input gate activates input only if $\widetilde{C}_t$'s input is represented by $U^g$, otherwise, it forgets the current input $x_t$ by not passing them to $h_t$. The extent of forgetting $x_t$ is increased when forget gate is activated (even) more by having a higher $f_t$. Assigning name of forget gate to $f_t$ is due to this.

**Huffman Coding and its relationship to proposed approach:** Huffman Coding is a Method for the Construction of Minimum-Redundancy Codes [11]. Huffman tree assigns shorter binary codes to more frequently occurring characters and longer codes to less frequent ones, thereby achieving compression. Figure 2 shows process of Huffman coding. A binary tree is generated from left to right taking the two least probable symbols and putting them together to form another equivalent symbol having a probability that equals the sum of the two symbols. The process is repeated until there is just one symbol. The tree can then be read backwards, from right to left, assigning different bits to different branches.

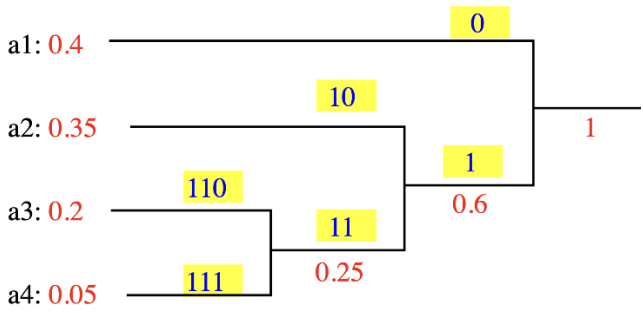

Figure 2. An example of Huffman coding. Credits:Wikipedia

**Shannon's Theorem and its relationship to proposed approach:** Shannon's theorem, also known as the noisy-channel coding theorem, fundamentally describes the maximum rate at which information can be reliably transmitted over a noisy communication channel. In essence, it establishes the theoretical limit for error-free data transfer through a channel, given a specific noise level.

This maximum rate is called the channel capacity (C) and is expressed by the following formula: C = B * log2(1 + S/N), where B is the bandwidth of the channel and S/N is the signal-to-noise ratio.

**Shannon Entropy and its relationship to proposed approach:** Shannon entropy, a concept from information theory, quantifies the average level of "information," "surprise," or "uncertainty" inherent in the possible outcomes of a random variable. It essentially measures how unpredictable a message or event is, with higher entropy indicating more uncertainty and lower entropy indicating more predictability.

**Markov Chain Classifiers and their relationship to proposed approach:** Markov chain classifiers are a type of classifier that uses Markov chains, a stochastic model, to classify data. They leverage the Markov property, where the future state depends only on the present state, not the past, to predict the probability of a sequence belonging to a particular class.

The simplification in the following formula removed all past effects on the next feature state. After this simplification, as 2 variables are remained, Formula 7 relationship is called 2Grams.

$$P(A_{n+1} = a \mid A_1 = a_1, A_2 = a_2 ..., A_n = a_n) = P(A_{n+1} = a \mid A_n = a_n) \quad (7)$$

### III. PROPOSED METHODS

*A. Input Reordering*

To grasp context from most part of a sentence, one has to ensure that input words are fed from the shallow levels of a binary hierarchy of indices to deeper levels (center of first and second sub-sentence half, with sub-sentence dissection scheme of two-equal-piece). Therefore, to avoid gradient vanishing (or exploding) from shallowest hierarchical sampling of tokens, they should be fed to LSTM sooner.

*1) Hierarchcial binary tree based LSTM*

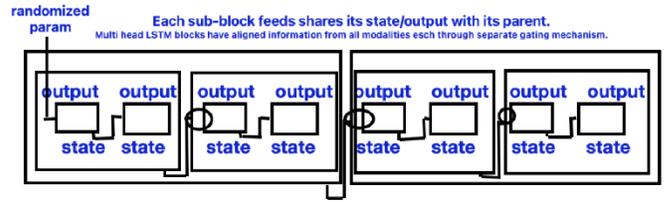

Figure 3. Initial idea to prioritize input tokens, and reduce gradient vanishing (and exploding)

Implementation of this approach using unit-change-approach is stopped, because changing inputs is way more simpler (for getting similar results and simualtions) than adding a unit (block switcher).

*2) Simple LSTM with reordered hierarchical binary tree based input*

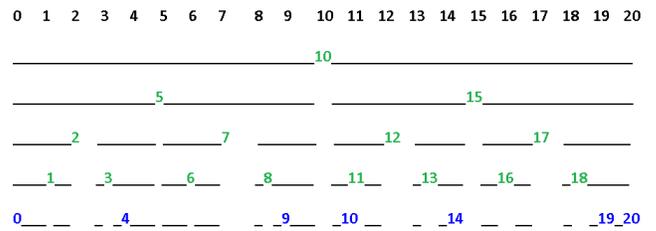

Figure 4. Hierarchical binary tree based reindexing. New indices orders: 10, 5, 15, 2, 7, 12, 17, 1, 3, 6, 8, 11, 13, 16, 18, 0, 4, 9, 10, 14, 19, 20. Indices' order here is reversed to make less indices with wider spread backpropagated sooner with less number of gradient chains in the Chain-Rule, , which directly affects the gradient vanishing. .

Pseudocode 1. Hierarchical binary tree based reindexing.

| Input: | TotalLength |
|---|---|
| Output: | orderSpaceList |
| Define function Pseudocode1( TotalLength) | Nitm <= TotalLength<br>Append(orderSpaceList, Nitm // 2)<br>Append(stackSpaceList, Nitm // 2) # center<br>Append(stackSpaceList0, 0) #start<br>Append(stackSpaceList1, Nitm) #end<br>While length(stackSpace) > 0:<br>  it = stackSpaceList.pop(0) # center<br>  it0 = stackSpaceList.pop(0) # start<br>  it1 = stackSpaceList.pop(0) # end<br>  if (it - it0)//2 >= 1 and (it1 - it) // 2 >= 1:<br>    Append(stackSpaceList, (it - it0) // 2 + it0 ) # center<br>    Append(stackSpaceList0, it0) #start<br>    Append(stackSpaceList1, it) #end<br>    #<br>    Append(stackSpaceList, (it1 - it) // 2 + it ) # center<br>    Append(stackSpaceList0, it) #start<br>    Append(stackSpaceList1, it1) #end<br>    #<br>    Append( orderSpaceList, (it1 - it) // 2  + it )<br>    Append( orderSpaceList, (it - it0) // 2  + it0 )<br>  else:<br>    # indices shown in dark blue in Figure 4<br>    a = set([0: Nitm]) - set(orderSpaceList)<br>    a = reverseList ( sorted (a) )<br>    Extend ( orderSpaceList, aaa )<br># To ensure wider span of tokens all over the sentence, we start by passing first layers to the LSTM unit.<br>orderSpaceList = reverseList ( orderSpaceList )<br>Return orderSpaceList |

*3) 2Gram Preservation and inspiration for reordered hierarchical binary tree based inputs*

Contiguous tokens have better interrelationships with each other. Therefore, we tested accuracy change due to using hierarchical binary reordering of 2-Gram pairs of indices. The results are promising (improving), and are shown in Section 4.

**Add hierarchy.** For n-Grams, we find hierarchy in Figure 4 as in Figure 5. Then we replace each number 'k' with n numbers ( i*k, i*k +1, … , i*k+ (k-1) ).

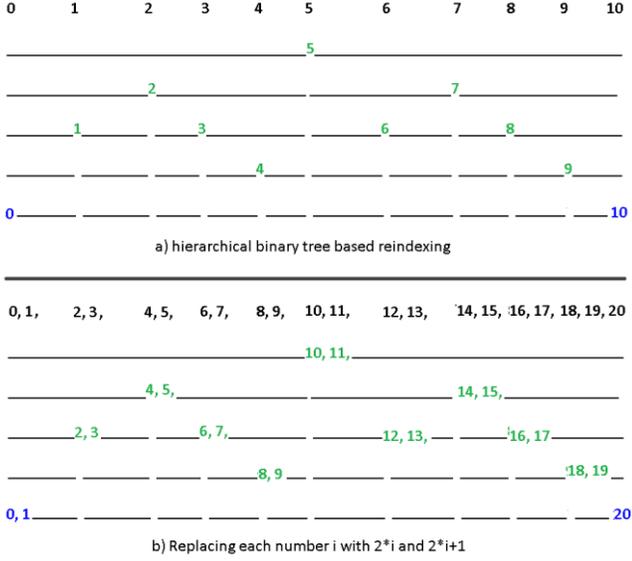

Figure 5. Hierarchical binary tree based reindexing for 10 tokens, plus number replacing. i is the index to change to 2 contiguous indices. New indices orders: 10, 11, 4, 5, 14, 15, 2, 3, 6, 7, 12, 13, 16, 17, 8, 9, 18, 19, 0, 1, 20.

By replacing each number n in Figure 5-a with 2*i and 2*i + 1, we come up with 2-Gram version of 20 number reindex, shown in Figure 5-b.

Indices' order here is reversed to make less indices with wider spread backpropagated sooner with less number of gradient chains in the Chain-Rule. Therefore the used indices order is: 20, 1, 0, 19, 18, 9, 8, 17, 16, 13, 12, 7, 6, 3, 2,15, 14, 5, 4, 11, 10.

Pseudocode 2. Hierarchical binary tree based reindexing for 10 tokens, plus number replacing with 2-Gram tuple.

| Input: | TotalLen, LenNGrams=2, orderSpaceList2= [] |
|---|---|
| Output: | orderSpaceList |
| Pseudocode | Nitm = TotalLength // LenNGrams<br><br>orderSpaceList= Pseudocode1(TotalLength = TotalLen//LenNGrams)<br><br>For o in orderSpaceList:<br>    Extend( orderSpaceList2, list(range(o, o - LenNGrams,-1))<br><br>Return orderSpaceList2 |

After creating the indices, it is necessary to start from the last index, moving backward to the first index. The reason is due to the behavior of LSTM. As the number of indices that are near the last index and close to each other are way higher than number of initial indices in the similar level of closeness, by beginning from starting index, LSTM will get caught into lots of nested nonlinearities that opt for gradient vanishing (or exploding), eventually making time series data with longer time steps tend to forget the previous data.

### B. Weights regularization

There are eight different parameters to optimize and regularize in LSTM: x based parameters ($U^i, U^f, U^o, U^g$), and h based parameters ($W^i, W^f, W^o, W^g$).

LSTM Parameters regularization have series of advantages:
- Using Lp norm ( $\|W\|_p = (|W_{1j}|^p + |W_{mj}|^p + |W_{2j}|^p + \ldots + |W_i*(j+m*I)|^p)^{\wedge}(\frac{1}{p})$ ) to find p for each of matrix parameters. I is number of rows and m is number of columns.
- Prevent from having parameters overfitting to training data by reducing the changeability of series parameters
- Optimize each matrix parameter with different multiplication weights

To regularize U and W series of parameters, the mentioned Lp norm formula has to be optimized (minimized). Therefore the following cost terms are added to the main cost function:

$$Cost_{terms} = \\
a_{i,1} * \|W^i\| p_{i,1} + \\
a_{f,1} * \|W^f\| p_{f,1} + \\
a_{o,1} * \|W^o\| p_{o,1} + \\
a_{g,1} * \|W^g\| p_{g,1} + \\
a_{i,2} * \|U^i\| p_{i,2} + \\
a_{f,2} * \|U^f\| p_{f,2} + \\
a_{o,2} * \|U^o\| p_{o,2} + \\
a_{g,2} * \|U^g\| p_{g,2} \quad (8)$$

*a* and *p* hyperparameter-vectors have to be optimized.

### C. Gate Nonlinearization

Figure 6 shows how to nonlinearize gates of LSTM. This proposed approach has the following advantages:
- New parameters easily increasees capacity of LSTM for deeper memorization capability.
- Parameters matrix multiplication followed by a ReLU activation nonlinearizes input values, provides more diverse gradient updates in the hands of LSTM learning procedure.

$$i_t = \sigma(x_t U^i + h_{t-1} W^i)$$
$$f_t = \sigma(x_t U^f + h_{t-1} W^f)$$
$$o_t = \sigma(x_t U^o + h_{t-1} W^o)$$
$$\tilde{C}_t = \tanh(x_t U^g + h_{t-1} W^g)$$

$$\Longrightarrow$$

$$i_t = \sigma(ReLU(x_t U^i)A^i + ReLU(h_{t-1} W^i)B^i)$$
$$f_t = \sigma(ReLU(x_t U^f)C^i + ReLU(h_{t-1} W^f)D^i)$$
$$o_t = \sigma(ReLU(x_t U^o)E^i + ReLU(h_{t-1} W^o)F^i)$$
$$\tilde{C}_t = \tanh(ReLU(x_t U^g)G^i + ReLU(h_{t-1} W^g)H^i)$$

Figure 6. Nonlinearizing LSTM gates with ReLU activation and new parameter matrices multiplication. Matrices from A to H are new m*m weight parameters.

## IV. RESULTS

### A. Hyperparameter Tuning

In this subsection, we provide the optimal hyperparameters we used to train LSTM. Number of epochs are set to 50, learning rate is selected as 0.01 from list of {0.1, 0.01, 0.001, 0.0001}. 100 is used for embedding dimension (variable length of each sample sentence). Hidden dimension is set to 50. Finally optimal sentence length is set to 32. Sentences lengths smaller than 32 are padded with 'UNK' tokens (32-length).

### B. Used Datasets

For first and second contributions (input reordering, weights regularization), dataset R8 is used, while for third idea (gate nonlinearization), dataset Amazon Fine Food Reviews (AFFR) is used. R8 is the subset of a collection of newswire articles from a financial newswire service, an essential benchmark for text categorization research. AFFR span a period of more than 10 years, including all ~500,000 reviews up to October 2012. Reviews include product and user information, ratings, and a plain text review.

10% of text is randomly selected from the training set to build validation set.

### C. Performance Comparisons of Each Proposed Method

In this subsection, comparison of accuracy between baseline (regular LSTM) versus three modifications in proposed LSTM training pipeline has been derived in Table 2. The results outperformed baseline algorithm, showing the promising-ness of the ideas.

Table 2. performance comparisons of each proposed method versus regular LSTM

|  | Training Accuracy | Testing Accuracy | Training Error | Testing Error |
|---|---|---|---|---|
| **Baseline (regular LSTM)** | 87.3 | 84.1 | 0.075 | 0.100 |
| **Input Reordering** | 88.9 | 85.0 | 0.069 | 0.094 |
| **Weight Regularization** | 87.5 | 84.5 | 0.042 | 0.100 |
| **Gate Nonlinearization** | 87.5 | 84.8 | 0.055 | 0.050 |

## V. CONCLUSION, FUTURE WORKS

In this work, 3 different solutions unnoticed in LSTM literature is implemented, i.e., binary tree based input reordering, weight regularization, and gate nonlinearization. The promising results of evaluations proved that the solutions are necessary for LSTM models to cure gradient vanishing, gradient exploding and low accuracy.

In this study, no comparison is conducted between Bi-LSTM and input reordering LSTM. Accuracy of bidirectional LSTM (b-LSTM) is not compared with proposed approach, which can also implemented with or without b-LSTM.